# On testing whether an Embedded Bayesian Network represents a probability model

Dan Geiger*
Technion
CS Dept
Israel, 32000

Azaria Paz†
Technion
CS Dept
Israel, 32000

Judea Pearl‡
UCLA
CS Dept
LA, CA, 90024

## Abstract

Testing the validity of probabilistic models containing unmeasured (hidden) variables is shown to be a hard task. We show that the task of testing whether models are structurally incompatible with the data at hand, requires an exponential number of independence evaluations, each of the form: "$X$ is conditionally independent of $Y$, given $Z$." In contrast, a linear number of such evaluations is required to test a standard Bayesian network (one per vertex). On the positive side, we show that if a network with hidden variables $G$ has a tree skeleton, checking whether $G$ represents a given probability model $P$ requires the polynomial number of such independence evaluations. Moreover, we provide an algorithm that efficiently constructs a tree-structured Bayesian network (with hidden variables) that represents $P$ if such a network exists, and further recognizes when such a network does not exist.

## 1 Introduction

Bayesian Networks possess several desirable properties for representing uncertain knowledge, especially in systems that perform diagnosis and forecasting. One such property is the ability to encode and update probabilistic knowledge economically, using modular representation and distributed computations. Another property is the ability to represent causal knowledge of the domain in a way that supports a wide variety of inferences, including prediction, abduction and the control of actions. A third property, which is the central topic of this paper, is the possibility of model validation, namely, testing by objective measurements whether a Bayesian network, constructed by a domain expert accurately represents the target domain. Much of the basic work on Bayesian networks can be found in [Pe88], and more recent advances are summarized in [Pe93a, Pe93b].

The appeal of the Bayesian network model has prompted researchers to suggest several useful extensions one of which is Embedded Bayesian Networks (e-BN) [PV91]. E-BNs enhance the language of Bayesian networks by allowing bidirected edges in addition to the directed edges permitted in Bayesian networks; each bidirected edge represents a pair of variables that are correlated but have no causal influence on each other. Such symmetrical correlations normally emanate from causal factors which the analyst chooses to exclude from formal analysis, either because they lie beyond the scope of the domain, or because they are inaccessible to direct measurement. In this paper, we address the task of validating Embedded Bayesian networks subject to the constraint that all causal factors responsible for the bidirected arcs are unobservable.

Model validation involves two subtasks; validating the qualitative graph structure provided by the expert and validating the numerical parameters associated with the edges of the graph. Each parameterized graph structure defines a joint probability distribution on the observed variables which, in principle, can be tested for compatibility with the observed data. A graph structure is said to be valid if it can be parameterized so as to define a probability distribution compatible with an observed distribution $P$. [1]

When the structure is valid, then iterative learning technique can be employed for tuning the parameters so as to fit a given stream of empirical observations [La91]. However, if the structure itself is erroneous, no parameter tuning can ever render the model compatible with the data. We, therefore focus our attention on the task of validating the structure of an embedded Bayesian network.

---

*Some of this work was done while the author visited Microsoft Research Center. Email: dang@cs.technion.ac.il.

†Some of this work was done while the author visited UCLA. Email: paz@cs.technion.ac.il.

‡E-mail: judea@cs.ucla.edu

[1] We assume that the entire distribution P is observed directly, which is a good approximation when the sample size is large.

Testing structural validity of an e-BN is much harder than that of a Bayesian network. In BN, the values of all parameters are uniquely determined from the observed distribution P by mere projection and, so, we can test for structural validity by checking whether the parameterized structure defines a probability distribution compatible with the observed data. Alternatively, a BN is known to be valid if and only if all the independencies implied by the BN hold in the observed distribution P and, moreover, this matching of independencies can be verified by testing a linear number of conditional independence statements (one per vertex).

Things are different in e-BNs. First, the set of parameters defining an e-BN model cannot be obtained by projection. Moreover, the domain of the hidden variables may be unbounded or undefined, so one cannot test model validity by first fitting parameters (say by maximum-likelihood techniques) and then testing the resulting model for agreement with P. Second, matching independencies is a necessary but not sufficient condition for e-BN validity. In other words, even if all independencies implied by the structure of a given e-BN hold in a distribution P, it is still possible that the e-BN is incapable of generating P (see [Verma and Pearl 1991] for a counterexample). Finally, while matching independencies could serve as a potentially quick way of screening invalid models, it is not clear how one would test the validity of all independencies in a given e-BN, because, in principle, the number of $(X, Y, Z)$ triplets to be tested is exponential with the number of vertices.

This last problem is the central theme of our paper to which we find a negative result. We prove that given an embedded Bayesian network $G$, checking whether $G$ faithfully represents independencies of an empirical distribution $P$, requires a number of evaluations of independence statements, each of the form: "$X$ is conditionally independent of $Y$, given that we observe $Z$", which grows exponentially in the number of vertices. This result is rather surprising since for Bayesian networks this task requires only a linear number of such evaluations. Obviously, to verify even one independence statement that contains $n$ variables requires to verify an exponential number of equalities, however, when the number of parents of each vertex is a fixed constant, then a linear number of statements can be verified by a linear number of hypothesis tests one per equality.

Notably, our negative result (Theorem 4) does not exclude the possibility that given additional properties of $P$ which are not shared by all probability models, a test of independence will be devised which requires only a polynomial number of independence evaluations. Consequently, it is advisable to consider probability distributions which are faithful [SV92] for which a test of independence may still be feasible.

On the positive side, we show that if $G$ has a tree skeleton, checking whether $G$ represents $P$ requires the evaluation of only polynomial number of independence assertions. Moreover, we provide an algorithm that efficiently constructs a tree-structured embedded Bayesian network that represents $P$ if such a network exists, and further recognizes when such a network does not exist.

## 2 Preliminaries

Let $U$ be a finite set of variables $\{u_1 \ldots u_n\}$ and let the domain of each $u_i$ be $d(u_i)$. A *probability model over* $U$ is a probability distribution of the form $P : d(u_1) \times \ldots \times d(u_n) \to [0, 1]$ where each $d(u_i)$, $i = 1, \ldots, n$, is a finite set. The class of probability models over $U$ is denoted by $\mathcal{P}$. A probability model over $U$ is *strictly-positive* if every combination of $U$'s values has a probability greater than zero. The class of strictly-positive probability models is denoted by $\mathcal{P}^+$. Note that $\mathcal{P}^+ \subseteq \mathcal{P}$.

An expression $I(X, Z, Y)$ where $X \neq \emptyset$, $Y \neq \emptyset$, and $Z$ (possibly empty) are disjoint subsets of a finite set $U$ is called an *independence statement*. A set of independence statements is called a *dependency model*. An independence statement $I(X, Z, Y)$ is said to *hold for* a probability model $P$ if for every value $\mathbf{X}$, $\mathbf{Y}$, and $\mathbf{Z}$ of $X$, $Y$, and $Z$, respectively,

$$P(\mathbf{X}, \mathbf{Y}, \mathbf{Z}) \cdot P(\mathbf{Z}) = P(\mathbf{X}, \mathbf{Z}) \cdot P(\mathbf{Y}, \mathbf{Z}). \quad (1)$$

The set of statements that hold for $P$ is denoted by $M(P)$ and is called the *dependency model induced by* $P$.

When $I(X, Z, Y) \in M(P)$, then $X$ and $Y$ are *conditionally independent* relative to $P$, and if in addition $Z = \emptyset$, then $X$ and $Y$ are *marginally independent* relative to $P$.

Eqs. (2) through (5) below are properties of conditional independence that hold for every probability model and Eq. (6) is a property that holds for every strictly positive probability model. Variants of these properties were first introduced by Dawid (1979) and further studied by Spohn (1980), Pearl and Paz (1985), Pearl (1988), and Geiger (1990).

Symmetry
$$I(X, Z, Y) \Rightarrow I(Y, Z, X) \quad (2)$$

Decomposition
$$I(X, Z, Y \cup W) \Rightarrow I(X, Z, Y) \quad (3)$$

Contraction
$$I(X, Z, Y) \,\&\, I(X, Z \cup Y, W) \Rightarrow I(X, Z, Y \cup W) \quad (4)$$

Weak-union
$$I(X, Z, Y \cup W) \Rightarrow I(X, Z \cup W, Y) \quad (5)$$

Intersection
$$I(X, Z \cup W, Y) \,\&\, I(X, Z \cup Y, W) \Rightarrow I(X, Z, Y \cup W) \quad (6)$$

The interpretation of each of these properties is straight forward. For example, Weak-union states that if $X$, $Y$, $Z$, and $W$ are sets of variables such that $I(X, Z, Y \cup W)$ is in $M(P)$ for some probability model $P$, then $I(X, Z \cup W, Y)$ must also be in $M(P)$. In other words, according to Weak-union, if $X$ and $Y \cup W$ are conditionally independent given $Z$ is known, then $X$ and $Y$ are conditionally independent given $Z \cup W$ is known.

Every dependency model that satisfies Eqs. (2) through (5) is called a *graphoid*. Hence, every probability model induces a graphoid. A Bayesian network and the corresponding d-separation criteria, defined below, is another example of a dependency model which is a graphoid. A graphoid defined by a Bayesian network usually serves to represent graphoids induced by probability models.

The primary advantage of a Bayesian network is that it allows a wide spectrum of independence assumptions to be conveniently considered by a model builder so that a practical balance can be established between computational needs and adequacy of conclusions. We now give a definition of a Bayesian network.

A *dag (directed acyclic graph)* is a directed graph with no parallel edges and no directed cycles. A *trail* in a dag is a subgraph whose underlying graph is a path in which no vertex appears twice. A vertex $b$ is called a *sink* on a trail $t$ if there exist two consecutive edges $a \to b$ and $b \leftarrow c$ on $t$. A trail $t$ is *active by a set of vertices* $Z$ if (1) every sink with respect to $t$ either is in $Z$ or has a descendant in $Z$ and (2) every other vertex along $t$ is outside $Z$. A trail is said to be *blocked* by $Z$ if it is not active by $Z$. An independence statement $I(X, Z, Y)$ holds in $D$ if $X$, $Y$, and $Z$ are disjoint subsets of vertices, and every trail between an element in $X$ and an element in $Y$ is blocked by $Z$. The set of all independence statements that hold in $D$ are denoted by $M(D)$. The set $M(D)$ is called the dependency model *induced* by $D$. A dag $D$ is an *I-map* of a probability model over $U$ if $I(X, Z, Y) \in M(D)$ implies $I(X, Z, Y) \in M(P)$. A dag $D$ is a *minimal I-map of* $P$ if whenever an edge is removed from $D$, the resulting dag is not an I-map of $P$. A dag $D = (U, E)$ is a **Bayesian network** of a probability model $P$ over $U$ if $D$ is a minimal I-map of $P$.

**Lemma 1 ([Pe88])** *A dependency model induced by a dag is a graphoid.*

Bayesian networks as defined above would have remained merely interesting mathematical objects unless an efficient procedure existed for testing whether a given dag is an I-map of a probability model. Since such a procedure does exist [Pe88], it became possible to construct a dag from causal knowledge and then test whether the dag constructed actually reflects reality as sensed by measured data.

A known example of a Bayesian network is given in Figure 1. It is depicted using the assertions that a

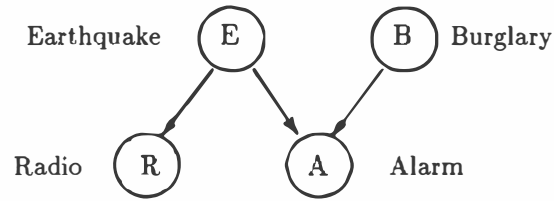

Figure 1

Burglary ($B$) and an Earthquake ($E$) may each activate an Alarm ($A$) and that Radio announcement ($R$) may follow an earthquake. Examining the dependency model induced by this dag we see, for example, that $B_r = \{I(B, \emptyset, E), I(R, E, \{A, B\})\}$ is a subset of $M(D)$. The first statement asserts that burglaries and earthquakes are independent. The second statement asserts that once an earthquake is known to happen or not to happen with certainty, whether or not a radio announcement is broadcasted is independent of both the alarm system and a burglary occurrence. Suppose $P(B, E, A, R)$ is a probability model describing the 16 possible outcomes, then one can test whether $I(B, \emptyset, E) \in M(P)$ and whether $I(R, E, \{A, B\})\} \subseteq M(P)$. If it is not the case, additional edges, such as an edge from $E$ to $B$, could be added to reflect reality more accurately.

Pearl and Verma [PV91] show that if $B_r \subseteq M(P)$, then $M(D) \subseteq M(P)$ and therefore $D$ is an I-map of $P$. In fact, each dag defines a linear-sized set like $B_r$, called a *recursive basis*, such that $B_r \subseteq M(P)$ implies $M(D) \subseteq M(P)$. Thus, I-mapness can be tested efficiently. Furthermore, Pearl and Verma show that $M(D)$ is precisely the set of statements that can be derived from a recursive basis by repeated application of Eqs. (2) through (5). For example, by applying, Decomposition, Symmetry, and Contraction one can derive $I(B, \emptyset, R) \in M(D)$ from $B_r$. It is worthy to mention that the definition of active trails cannot be enhanced because every statement entailed from a recursive basis can be derived using Eqs. (2) through (5) [GVP90].

## 3  Embedded Bayesian Nets

In real-life applications each edge in a Bayesian network is usually directed from cause to effect. However, two variables are often correlated without having a causal influence of one on the other. Such a situation arises when a latent variable, not modeled by the network, is a common cause of the two observable variables. Embedded Bayesian networks, defined below, encode such a situation with a bidirected edge, $x \leftrightarrow y$. Each such edge is equivalent to a trail $x \leftarrow \alpha \to y$ in a Bayesian network in which the variable $\alpha$ is latent (unobservable). For example, two symptoms $x$ and $y$ of a disease $\alpha$ can be modeled by the network $x \leftrightarrow y$.

An *embedded directed graph* is a graph with two types of edges: directed edges, $x \to y$, and bidirected edges, $x \leftrightarrow y$. A path $a_1, \ldots, a_n$ is *directed from $a_1$ to $a_n$* if each edge $(a_i, a_{i+1})$ is a directed edge from $a_i$ to $a_{i+1}$. A *directed cycle* is a *directed path* where $a_1 = a_n$. An *E-dag* is an embedded directed graph that has no parallel edges and no directed cycles. A *descendant* $y$ of a vertex $x$ is any vertex such that there exists a directed path from $x$ to $y$. A *trail* in a dag is a subgraph whose underlying graph is a path in which no vertex appears twice. A vertex $b$ is called a *sink* on a trail $t$ if there exist two consecutive edges $(a, b)$ and $(b, c)$ on $t$ such that none of these two edges is a directed edge that points away from $b$. (If $b$ is incident with only one edge on $t$, then $b$ is not a sink). A trail $t$ is *active by a set of vertices $Z$* if (1) every sink on $t$ either is in $Z$ or has a descendant in $Z$ and (2) every other vertex along $t$ is outside $Z$. A trail is said to be *blocked* by $Z$ if it is not active by $Z$.

An independence statement $I(X, Z, Y)$ holds in an E-dag $G$ if $X$, $Y$, and $Z$ are disjoint subsets of vertices, and every trail between an element in $X$ and an element in $Y$ is blocked by $Z$. The set of independence statements that hold in an E-dag $G$ are denoted by $M(G)$. The set $M(G)$ is called the dependency model *induced* by $G$. An E-dag $G$ is an *I-map* of a probability model over $U$ if $I(X, Z, Y) \in M(D)$ implies $I(X, Z, Y) \in M(P)$. An E-dag $G$ is a *minimal I-map of $P$* if whenever an edge is removed from $G$, the resulting E-dag is not an I-map of $P$. An E-dag $G = (U, E)$ is an *Embedded Bayesian network of a probability model $P$ over $U$* if $G$ is a minimal I-map of $P$.

Similar to dags, we have:

**Lemma 2** *A dependency model induced by an E-dag is a graphoid.*

**Proof.** Let $D$ be an E-dag and let $D'$ be a dag formed from $D$ by replacing each bidirected edge $x \leftrightarrow y$ with a new vertex $\alpha$ and two directed edges $\alpha \to x$ and $\alpha \to y$. It is easy to verify that whenever $X$, $Y$ and $Z$ are disjoint sets of vertices of $D$, then $I(X, Z, Y) \in M(D)$ if and only if $I(X, Z, Y) \in D'$. The lemma now follows directly from Lemma 1. □

In order to check whether $G$ is an I-map of $P$ one could naively test that each element of $M(G)$ is in $M(P)$. We call each such test a *membership test*. Analogously to what we have claimed in the previous section for dags, one can check whether an E-dag $G$ is an I-map of $P$ using less than $|M(G)|$ membership tests. The rest of this section shows how.

A set of statements $\Sigma$ *entails* a set of statements $\Gamma$ if for every probability model $P$, $\Sigma \subseteq M(P)$ implies $\Gamma \subseteq M(P)$. The set $\Sigma$ *positively entails* $\Gamma$ if for every strictly-positive probability model $P$, $\Sigma \subseteq M(P)$ implies $\Gamma \subseteq M(P)$. A set $B \subseteq M(G)$ is called a *probabilistic basis* of an E-dag $G$ if $B$ entails $M(G)$. The set $B$ is a *minimum* probabilistic basis if there exists no other probabilistic basis $B'$ of $G$ satisfying $|B'| < |B|$.

**Lemma 3** *Let $B$ be a basis of an E-dag $G$ and let $P$ be a probability model. If $B \subseteq M(P)$, then $M(G) \subseteq M(P)$.*

**Proof.** The proof follows from the definition of entailment where $\Sigma = B$ and $\Gamma = M(G)$. □

**Lemma 4** *Let $B$ be a minimum basis of an E-dag $G$ and let $P$ be a probability model. The number of membership tests needed in order to determine whether $M(G) \subseteq M(P)$ is equal to $|B|$.*

**Proof.** Let $k$ be the smallest number of membership tests needed in order to determine whether $M(G) \subseteq M(P)$. First we show that $k \leq |B|$. If $B \subseteq M(P)$, then $M(G) \subseteq M(P)$ because $B$ is a probabilistic basis of $G$ (Lemma 3). Testing whether $B \subseteq M(P)$ requires at most $|B|$ membership tests—one for each statement in $B$. Thus $k \leq |B|$. Next we show that $k \geq |B|$ which implies the claim.

Let $R \subseteq M(G)$ be a smallest set of independence statements upon which the assertion $M(G) \subseteq M(P)$ is made. By the definition of $k$, $|R| = k$. Furthermore, $R$ is a probabilistic basis of $G$ because otherwise $M(G) \subseteq M(P)$ cannot have been asserted. Since $|B|$ is a minimum basis, $k \geq |B|$. □

We note that a recursive basis of a dag is in fact a probabilistic basis of at most $n$ statements—one per vertex in $G$. Hence, according to Lemma 4, testing whether a dag is an I-map of a probability model $P$ requires only $n$ membership tests. This property of having a poly-sized basis, as we show in the next section, does not extend to E-dags, and thus, the number of membership tests needed in order to check whether an E-dag $G$ is an I-map of some probability model $P$ may grow exponentially in the number of vertices of $G$.

This result is rather discouraging because it says that if the number of variables is large enough and some variables are unobservable, then one cannot verify in reasonable time that a given causal description is correct by simply observing the world. In Section 5 we show that some E-dags can be verified using a polynomial number of membership tests.

To prove the non-existence of a poly-sized basis in some E-dags we need the following completeness theorem concerning independence statements of the form $I(X, \emptyset, Y)$ which we call *marginal statements*.

**Theorem 1 ([GPP91])** *Let $B$ be a set of marginal statements and let $B^*$ be the set of all marginal statements that $B$ entails. Then $\sigma \in B^*$ if and only if $\sigma$ can be derived from $B$ by repeated applications of the following properties:*

*M-symmetry*
$$I(X, \emptyset, Y) \Rightarrow I(Y, \emptyset, X) \qquad (7)$$

*M-decomposition*

$$I(X, \emptyset, Y \cup W) \Rightarrow I(X, \emptyset, Y) \qquad (8)$$

*M-mixing*

$$I(X, \emptyset, Y) \& I(X \cup Y, \emptyset, Z) \Rightarrow I(X, \emptyset, Y \cup Z) \qquad (9)$$

This theorem states that the above three properties, which are satisfied by any graphoid (M-Mixing is derived from Weak-union and Contraction), are sound and complete for the set of probability models $\mathcal{P}$; Every marginal statement that is entailed can be derived and vice versa. Theorem 1 remains correct when $B^*$ is redefined to be the set of statements positively entailed by $B$. Thus, strictly-positive probability models do not share any property for marginal statements that is not already shared by all probability models.

## 4 Verifying Embedded Bayesian Nets

We now construct a sequence of E-dags, $G_k$, $k \geq 1$, such that each $G_k$ has $2k + 2$ vertices and the cardinality of any of $G_k$'s probabilistic bases is larger or equal to $2^k$. Consequently, due to Lemma 4, testing whether $G_k$ is an I-map of a given probability model $P$ requires a number of membership tests that grows exponentially in the number of vertices.

We define $G_k$ as follows. The vertices of $G_k$ are $C \cup D$ where $C = \{c_0, c_1, \ldots, c_k\}$ and $D = \{d_0, d_1, \ldots, d_k\}$. All edges in $G_k$ are bidirected. All vertices in $C$ are connected with a bidirected edge to each other and all vertices in $D$ are connected with a bidirected edge to each other. That is, $C$ and $D$ are cliques. For $i = 1 \ldots k$ (but not for i=0), $c_i$ and $d_i$ are connected with a bidirected edge.

**Theorem 2** *If $I(X, Z, Y) \in M(G_k)$, then there exists a partition $Z'$, $Z''$ of $Z$, such that $I(X \cup Z', \emptyset, Y \cup Z'') \in M(G_k)$.*

**Proof.** Suppose $\sigma = I(X, Z, Y) \in M(G_k)$. The statement $\sigma$ has the form $I(X, C' \cup D', Y)$ where $C' \subseteq C$, and $D' \subseteq D$. It cannot be the case that both $X \cap C \neq \emptyset$ and $Y \cap C \neq \emptyset$ because any two vertices in $C$ are connected with an edge. Similarly, it cannot be the case that both $X \cap D \neq \emptyset$ and $Y \cap D \neq \emptyset$. Suppose now, with out loss of generality, that $X \cap D = \emptyset$. Hence, since $X$ is not the empty set, $X \cap C \neq \emptyset$, and therefore, $Y \cap C = \emptyset$. So $\sigma$ has the form $I(C'', C' \cup D', D'')$ where $C'' \subseteq C$ and $D' \subseteq D$. The set $C' \cup D'$ contains no pair $(c_i, d_i)$, $i \neq 0$, because otherwise each element $x \in C''$ would be connected to each element $y \in D''$ via the active trail $x \leftrightarrow c_i \leftrightarrow d_i \leftrightarrow y$. Similarly, the set $C'' \cup D'$ contains no pair $(c_i, d_i)$, $i \neq 0$, because otherwise $c_i \in C''$ would be connected to each element $y \in D''$ via the active trail $c_i \leftrightarrow d_i \leftrightarrow y$. Thus, $I(\{c\}, \emptyset, \{d\}) \in M(G_k)$ for every $c \in C' \cup C''$ and for every $d \in D'$. Hence, $I(C' \cup C'', \emptyset, D') \in M(G_k)$. Symmetric arguments imply $I(D' \cup D'', \emptyset, C') \in M(G_k)$. The last two statements together with $I(C'', C' \cup D', D'') \in M(G_k)$ derive $I(C' \cup C'', \emptyset, D' \cup D'') \in M(G_k)$, using Symmetry, Weak-union and Contraction, and so, $\{C', D'\}$ is the desired partition of $Z$. $\square$

**Theorem 3** *Every probabilistic basis $B$ of $G_k$ satisfies $|B| \geq 2^k$.*

**Proof.** First we show that it is sufficient to prove this theorem for probabilistic bases that consist solely of marginal statements.

Suppose $B$ is a probabilistic basis of $G_k$ and let $\sigma = I(X, Z, Y)$ be a statement in $B$. According to Theorem 2, there exists a partition $Z', Z''$ of $Z$, such that $\sigma_m = I(X \cup Z', \emptyset, Y \cup Z'')$ is in $M(G_k)$. Note that $\sigma_m$ entails $\sigma$ due to Symmetry and Weak-union. Let $B_m$ be a set of marginal statements obtained by replacing each statement $\sigma$ in $B$ with a statement $\sigma_m$ of the form just defined. It follows that $B_m$ entails $B$ and that $|B_m| = |B|$. Consequently, $B_m$ is a basis that has the same size of $B$. Hence, if every probabilistic basis consisting solely of marginal statements satisfy $|B| > k$, then every probabilistic basis satisfies this inequality as well.

Next we define a set $T$ of marginal statements of size $2^k$ and show that every probabilistic basis that consists solely of marginal statements must have a size larger than $|T|$. Recall that the vertices in $G_k$ are $C \cup D$. Let $T$ be the set of all marginal statements having the form $I(\{c_0\} \cup C', \emptyset, \{d_0\} \cup D')$ where $C' \subseteq C \setminus \{c_0\}$, $D' \subseteq D \setminus \{d_0\}$, and $C' \cup D'$ contains precisely one vertex of each pair $\{c_i, d_i\}$, $i = 1 \ldots k$. Note that the cardinality of $T$ is indeed $2^k$ and that $T \subseteq M(G_k)$ because each trail between a vertex in $\{c_0\} \cup C'$ and a vertex in $\{d_0\} \cup D'$ contains a vertex that is a sink on that trail.

Let $B$ be a probabilistic basis of $G_k$ that consists solely of marginal statements. Since $T \subseteq M(G_k)$ and $B$ is a basis, it follows that $B$ entails $T$. Due to the completeness theorem 1, $B$ entails $T$ if and only if for each $\sigma \in T$, there exists a derivation of $\sigma$ from $B$ using M-symmetry, M-decomposition and M-mixing. Let $\sigma$ be any statement in $T$ such that $\sigma$ has a derivation chain from $B$ of length $l$. Define the *symmetric image of a statement* $I(X, \emptyset, Y)$ to be the statement $I(Y, \emptyset, X)$. We now prove by induction on $l$ that either $\sigma \in B$ or the symmetric image $sym(\sigma)$ of $\sigma$ is in $B$. Consequently, $|B| \geq |T| \geq 2^k$ which proves the Theorem.

The case $l = 0$ is trivial; a derivation chain of length zero means that $\sigma \in B$. Assume the inductive claim holds for $l \leq s$. Let $\sigma \in T$ be a statement whose shortest derivation chain has a length of $s + 1$. At the last derivation step either M-symmetry, M-decomposition or M-mixing where used. We consider each case separately. If $\sigma$ is derived from $\sigma'$ by Symmetry, then $\sigma'$ is the symmetric image of $\sigma$. By the induction hypothesis, either $\sigma'$ or $sym(\sigma')$ is in $B$. Hence, either $\sigma$ or $sym(\sigma)$ is in $B$ as well. Next suppose $\sigma$ is derived by decomposition from $\sigma'$. If $\sigma'$ has the same number of vertices as $\sigma$, then $\sigma$ and $\sigma'$ are the same statements and therefore by the induction hypothesis, either $\sigma'$ or $sym(\sigma')$ is in $B$. Hence, either $\sigma$ or $sym(\sigma)$ is in $B$ as well. However, the number of vertices in $\sigma'$ cannot be strictly greater than that in $\sigma$; Assume it were greater. Then $\sigma'$ is not in $M(G)$ because the addition of any vertex to a statement in $T$ creates an active trail since for some $i$, $c_i$ would belong to $C'$ and $d_i$ would belong to $D'$. But $\sigma'$ is derived from a probabilistic basis of $G_k$ using properties that hold in every E-dag, and therefore, $\sigma'$ must be in $M(G)$. Contradiction.

Finally, suppose $\sigma = I(\{c_0\} \cup C', \emptyset, \{d_0\} \cup D')$ is derived using Mixing from two previous statements in the derivation chain $\gamma_1 = I(U, \emptyset, V)$ and $\gamma_2 = I(U \cup V, \emptyset, W)$. Both $\gamma_1$ and $\gamma_2$ must be in $M(G_k)$ because they were derived from a probabilistic basis of $G_k$ using properties that hold in every E-dag. If M-mixing is applied, the resulting statement has the form $I(U, \emptyset, V \cup W)$ where $V \cup W$ equals $\{d_0\} \cup D'$. Thus, either $V$ or $W$ must be empty or else $I(U \cup V, \emptyset, W) \notin M(G_k)$. Consequently, either $\gamma_1 = \sigma$ or $\gamma_2 = \sigma$. Thus, by the induction hypothesis, $\sigma$ or $sym(\sigma)$ is in $B$. □

We have thus derived the main claim of this section.

**Theorem 4** *Testing whether an E-dag $G$ is an I-map of a probability model $P$ requires, in the worst case, a number of membership tests that grows exponentially in the number of vertices of $G$.*

**Proof** Consider the E-dags $G_k$, $k \geq 1$. According to Theorem 3, each basis $B$ of $G_k$, in particular a minimum basis, satisfies $|B| \geq 2^k$ where $2k+2$ is the number of vertices in $G_k$. Lemma 4 shows that $|B|$ is the required number of membership tests, thus proving the claim. □

## 5  Verifying Embedded Bayesian Trees

An *E-tree* is an E-dag whose underlying graph is a tree. In this section we show that each E-tree has a probabilistic basis of a polynomial size and that such a basis can be easily found. Hence according to Lemma 4 testing whether an E-tree is an I-map of a probability model $P$ can be done using polynomial number of membership tests.

Let $T = (U, E)$ be an arbitrary E-tree. Let $x$ be a vertex in $T$. Let $s_1, \ldots, s_l$ be the set of vertices such that there exists a directed edge from each $s_i$ to $x$. Let $q_1, \ldots, q_k$ be all other vertices in $T$ which are connected to $x$ with an edge. Let $S_i$ be the set of vertices such that for each $y \in S_i$ the single trail connecting $y$ with $x$ passes through $s_i$. Similarly, let $Q_i$ be the set of vertices such that for each $y \in Q_i$ the single trail connecting $y$ with $x$ passes through $q_i$.

The set of all vertices in $T$ is denoted by $U$. For each vertex $x$ in $T$ and for each $s_i$ let $\sigma_i^x = I(S_i, \{x\}, U \setminus S_i \setminus \{x\})$. For each $q_i$, let $\gamma_i^x = I(Q_i, \emptyset, R_i^x)$ where $R_i^x$ is set of all vertices that are not descendants of $x$ and are not in $Q_i$. Let $B_T$ be the union over all vertices $x$ in $T$ of the set $\{\sigma_i^x, \gamma_j^x \mid i = 1\ldots l, j = 1\ldots k\}$. Clearly the size of $B_T$ is no greater than $n^2$ where $n$ is the number of vertices in $T$ because $l + k$, the number of edges adjacent with $x$, is less or equal to $n$. Furthermore, $B_T \subseteq M(T)$.

The rest of this section shows that $B_T$ is a probabilistic basis of $T$. Consequently, in order to test whether $T$ is an I-map of some probability model $P$ one must merely test that $B_T \subseteq M(P)$ using at most $n^2$ membership tests.

A statement $I(X, Y, Z)$ is *simple* if $X$ and $Y$ are singletons. The set of independence statements that can be derived from a statement $\sigma$ using Weak-union and Decomposition is denoted by $A(\sigma)$. (Note that $\sigma \in A(\sigma)$). The set of all simple statements in $A(\sigma)$ is denoted by $A_s(\sigma)$.

**Lemma 5** *Let $\sigma$ be an independence statement and $P$ be a probability model. If $A_s(\sigma) \subseteq M(P)$, then $\sigma \in M(P)$.*

**Proof.** Let $\sigma = I(X, Z, Y)$ and let $\sigma' = I(X', Z', Y')$ be an element of $A(\sigma)$. We prove by induction on the size of $X' \cup Y'$ that if $A_s(\sigma) \subseteq M(P)$, then $\sigma' \in M(P)$. Since $\sigma$ is in $A(\sigma)$, it follows from this induction that $\sigma \in M(P)$. Induction basis: If $|X' \cup Y'| = 2$, then $\sigma'$ is in $A_s(\sigma)$ and hence in $M(P)$ as well. Induction step: Assume that all $\sigma'$ in $A_s(\sigma)$ with $|X' \cup Y'| \leq k$ are in $M(P)$ and let $\sigma'' = I(X'', Z'', Y'')$ be a statement in $A_s(\sigma)$ with $Y'' = Y' \cup \{a\}$, $a$ is a singleton and $|X'' \cup Y''| = k+1$. Consider the statements $\sigma_1'' = I(X'', Z'' \cup \{a\}, Y')$ and $\sigma_2'' = I(X'', Z'', \{a\})$. Both $\sigma_1''$ and $\sigma_2''$ can be derived by Weak-union and Decomposition from $\sigma''$ which is either equal to $\sigma$ or can be derived from $\sigma$ by Weak-union and Decomposition. Thus $\sigma_1''$ and $\sigma_2''$ can be derived from $\sigma$ by Weak-union and Decomposition and are therefore in $A(\sigma)$. But $|X'' \cup \{a\}| \leq k$ and $|X'' \cup Y'| \leq k$ so that, by the induction hypothesis, both $\sigma_1''$ and $\sigma_2''$ are in $M(P)$. Finally $\sigma''$ can be derived from $\sigma_1''$ and $\sigma_2''$ by Contraction and $M(P)$ is closed under Contraction thus implying that $\sigma''$ is in $M(P)$. A similar argument proves the symmetric case when $X'' = X' \cup \{a\}$ and $|X'' \cup Y''| = k+1$. □

**Lemma 6** *Let $D$ be an E-DAG and let $P$ be a probability model. If all the simple statements in $M(D)$ are in $M(P)$ then $M(D) \subseteq M(P)$.*

**Proof.** If $\sigma \in M(D)$, then $A_s(\sigma) \in M(D)$ because, according to Lemma 2, $M(D)$ is closed under Decomposition and Weak-union. If all the simple statements in $M(D)$ are in $M(P)$, then in particular $A_s(\sigma) \in M(P)$. Hence, by Lemma 5, $\sigma \in M(P)$. Thus, $M(D) \subseteq M(P)$. □

The next theorem shows that $B_T$ is a probabilistic basis of $T$.

**Theorem 5** *Let $T$ be an E-tree and $P$ be a probability model. Then, $B_T \subseteq M(P)$ implies $M(T) \subseteq M(P)$.*

**Proof.** Let $\sigma = I(\{a\}, Z, \{b\})$ be an arbitrary simple statement in $M(T)$. We will show that if $B_T \subseteq M(P)$, then $I(\{a\}, Z, \{b\}) \in M(P)$. Consequently, due to Lemma 6, $M(T) \subseteq M(P)$ which is what we need to show. Since $T$ is a tree, there is a unique trail $t$ in $T$ connecting $a$ with $b$. Since this trail is blocked by $Z$, there are two cases to consider. Either (1) some sink $x$ on $t$ and all $x$'s descendants are not in $Z$, or (2) some vertex $x$ that is not a sink on $t$ is in $Z$.

If the first case occurs then consider the independence statement $\sigma' = I(Q_i, \emptyset, R_i^x) \in B_T$ where $a \in Q_i$ and $b \in R_i^x$. Such a statement exists in $B_T$ according to $B_T$'s definition. Furthermore, $R_i^x$ does not contain $x$ or any of $x$'s descendants and nor does $Q_i$. Thus, $Z \subseteq Q_i \cup R_i^x$. The statement $\sigma$ is therefore derivable from $\sigma'$ by Symmetry, Decomposition, and Weak-union which hold for $M(P)$. Hence, if $B_T \subseteq M(T)$, then $\sigma$ is in $M(P)$.

If the second case occurs then consider the independence statement $\sigma' = I(S_i, \{x\}, U \setminus S_i \setminus \{x\})$ which by definition is in $B_T$ where $a \in S_i$ and $b \in U \setminus S_i \setminus \{x\}$. Recall that all vertices of $T$ appear in $\sigma'$. Thus, if $B_T \subseteq M(P)$, then $\sigma \in M(P)$ since $\sigma$ can be derived from $\sigma'$ by Symmetry, Decomposition and Weak-union which hold in $M(P)$. □

It is worthy to note that one can define for each E-tree another polynomial basis $B_s$ that has the same number of independence statements as $B_T$ but some statements include less vertices. $B_s$ is defined as follows. For each $x$ in $T$, let $B_s(x)$ be the set $\{I(S_i, \{x\}, U \setminus S_i \setminus \{x\}) \mid i = 1, \ldots, l\}$ (as in $B_T$), and let $B'_s(x)$ be the set $\{I(Q_i, \{x\}, \bigcup_{j \neq i} Q_j) \mid i = 1, \ldots, k\}$. Let $B_s$ be the union over all vertices $x$ of $T$ of the set $B_s(x) \cup B'_s(x)$. The set $B_s$ is a probabilistic basis of $T$ because $B_s$ entails $B_T$ using Eqs. (7) through (9). In fact, every marginal statement in $M(T)$ can be derived from $B_s$ using these three properties. The proofs of these claims are omitted.

## 6  Learning Embedded Bayesian Trees

In section 5 we have analyzed the task of testing whether a given E-tree is an I-map of a given probability model $P$. Now we are concerned with the much more complicated task of synthesizing an E-tree that represents a given probability model, if such an E-tree exists, and recognizing when one does not exist. To facilitate our investigation we make two assumptions. First we consider only strictly positive probability models. This assumption is justified whenever categorical relationships can be excluded from the representation (as often happens, for example, in medical domains). Second, we only search for E-trees that represent $P$ well, as defined below.

A trail $t$ is called a *trek* if no vertex of $t$ is a sink on $t$. An E-dag $D$ that is a minimal I-map of a probability model $P$ is said to **represent $P$ well** if whenever two vertices $a$ and $b$ are connected with a trek in $D$, then $a$ and $b$ are marginally dependent, i.e., $I(\{a\}, \emptyset, \{b\})$ does not hold in $P$. Equivalently, we will say that $P$ is *well-represented* by $D$.

The assumption of well-representation is quite natural because one expects that changes in a variable $a$ on one side of a trek will reflect through the trek towards $b$ on the other end of the trek, thus making the two variables dependent.

The algorithm below determines whether a given strictly positive probability model $P$ can be well-represented by an E-tree and it finds such an E-tree if one exists. This result generalizes our claims in [GPP90, GPP93] in the sense that we now deal with embedded Bayesian networks instead of just Bayesian networks. There are examples in which the algorithm below recovers an E-tree I-map while our previous algorithm fail to recover a tree I-map because none exists. (For example, a chain of three bidirected edges).

## The Recovery Algorithm

**Input:** A strictly-positive probability model $P$ over $U$

**Output:** An E-tree that represents $P$ well if such exists, or acknowledgment that no such network exists.

1. Start with a complete undirected graph having $U$ as its vertex set.

2. Remove every edge $a - b$ for which $I(\{a\}, U \setminus \{a, b\}, \{b\})$ holds in $P$.

3. Remove every edge $a - b$ for which $I(\{a\}, \emptyset, \{b\})$ holds in $P$.

4. Let $R_G$ be the resulting graph. If $R_G$ is not a tree, then "FAIL".

5. Orient every pair of edges $(a, b)$ and $(b, c)$ towards $b$ whenever $I(\{a\}, \emptyset, \{c\})$ holds in $P$. (Note that each edge can be oriented twice–once to each direction).

6. Orient the remaining edges without introducing new sinks on any trail.

7. If the resulting E-tree does not represent $P$ well then "FAIL". Otherwise, output the resulting network.

Step 7 is done using polynomial number of independence statements as shown in Section 5.

The following claims establish the correctness of the algorithm. First we define a *skeleton* of an E-tree $T$ to be the underlying undirected graph of $T$.

**Theorem 6** *Let $P$ be a strictly-positive probability model. If $P$ can be well-represented by an E-tree $T$, then the skeleton of $T$ is equal to $R_G$—the graph constructed in step 3.*

Theorem 6 shows that step 3 of the algorithm identifies the skeleton of an E-tree that represents $P$ well, if such exists. Thus, if $P$ can be well-represented by an E-tree, then it must be one of the orientations of the undirected graph $R_G$ produced by step 3. Hence by checking all possible orientations of this graph, one can decide whether a strictly-positive model can be well-represented by an E-tree. Consequently, all E-trees that represent a strictly positive model $P$ must have the same skeleton.

The next theorem justifies an efficient way of establishing some orientations of the skeleton of $R_G$.

**Theorem 7** *Let $P$ be a strictly-positive probability model. If $T$ is an E-tree that represents $P$ well, and $a - b - c$ is a chain in the skeleton of $T$, then $b$ is a sink on $a - b - c$ if and only if $I(\{a\}, \emptyset, \{c\})$ holds in $P$.*

**Proof.** If $b$ is a sink on $a - b - c$, then $I(\{a\}, \emptyset, \{c\})$ is in $M(T)$ and is therefore in $M(P)$. Otherwise $a$ and $c$ are connected by a trek in $T$ which implies, by the well-representation assumption, that $I(\{a\}, \emptyset, \{c\})$ is not in $M(P)$. □

Step 6 leaves us freedom to choose the orientation of some edges in the skeleton. For example, the E-trees: $a \leftrightarrow b \rightarrow c$, $a \leftarrow b \leftrightarrow c$, and $a \leftarrow b \rightarrow c$ are three possible orientations of $a - b - c$. However, these three E-trees are indistinguishable (isomorphic) in the sense that they induce the same dependency models. Hence no algorithm that relies on independence statements can distinguish between them. On the other hand, the E-trees $a \leftrightarrow b \leftarrow c$ and $a \leftrightarrow b \leftrightarrow c$, are distinguishable from the previous three E-trees because both portray a new independence assertion, $I(\{a\}, \emptyset, \{c\})$, which is not represented in either of the former three E-trees. Our algorithm uses this distinction to orient these edges.

Two Embedded Bayesian networks $D_1$ and $D_2$ are *isomorphic* if $M(D_1) = M(D_2)$. Isomorphism defines the theoretical limitation on the ability to identify directionality of edges using information about independence.

**Theorem 8** *Two E-trees $T_1$ and $T_2$ are isomorphic iff they share the same skeleton and each of their corresponding trails have the same sinks on them.*

**Sufficiency:** If $T_1$ and $T_2$ share the same skeleton and have the same sinks on their corresponding trails then every active trail in $T_1$ is an active trail in $T_2$ and vice versa. Thus, $M(T_1)$ and $M(T_2)$, the dependency models corresponding to $T_1$ and $T_2$ respectively, are equal.

**Necessity:** $T_1$ and $T_2$ must have the same set of vertices $U$, for otherwise their dependency models are not equal. If $a \rightarrow b$ is an edge in $T_1$ and not in $T_2$, then the statement $I(\{a\}, U \setminus \{a, b\}, \{b\})$ is in $M(T_1)$ but not in $M(T_2)$. Thus, if $M(T_1)$ and $M(T_2)$ are equal, then $T_1$ and $T_2$ must have the same skeleton. Assume $T_1$ and $T_2$ have the same skeleton and that $a - c - b$ is a trail in these trees but that $c$ is a sink on that trail in $T_1$ while not being a sink on that trail in $T_2$. The trail $a - c - b$ is the only trail connecting $a$ and $b$ in $T_2$ because $T_2$ is an E-tree and it has the same skeleton as $T_1$. Since $c$ is not a sink on this trail in $T_2$, $I(\{a\}, \{c\}, \{b\}) \in M(T_2)$. However, $I(\{a\}, \{c\}, \{b\}) \notin M(T_1)$ because the trail $a \rightarrow c \leftarrow b$ is active by $\{c\}$. Thus, if $M(T_1)$ and $M(T_2)$ are equal, then $T_1$ and $T_2$ must have the same sinks on the each of the corresponding trails. □

Theorem 8 shows that all orientations of step 6 that do not introduce a new sink on any trail yield isomorphic E-trees because these E-trees satisfy the requirement of the theorem. Thus, in order to decide whether or not $P$ can be well-represented by an E-tree it is sufficient to examine **one** E-tree produced by step 6, as performed by step 7, because all other E-trees are isomorphic.

Note that it is not only sufficient but actually necessary to examine one E-tree produced by step 6 for I-mapness because there are cases where no orientation of $R_G$ yields an I-map, as shown by the following example. Let $M(P) = \{I(\{x_1\}, \emptyset, \{x_2\}), I(\{x_1\}, \emptyset, \{x_3\}) +$ *symmetric images*$\}$. Then the undirected graph $R_G$ consists of three vertices $\{x_1, x_2, x_3\}$ and one undirected edge $(x_2, x_3)$ but none of its three orientations yields an I-map of $M(P)$. Furthermore, note that Theorem 8 cannot be extended from E-trees to E-dags (although it does extend to every dag) because there are examples where two E-dags which have different skeletons are both minimal I-maps of the same dependency model $M(P)$.

## 7 Discussion

This paper exposes some of the basic difficulties in testing the validity of probabilistic models containing unmeasured (hidden) variables. On one hand, such models allow greater freedom in parameterizing the hidden links, requiring only that the chosen parameters be compatible with the observed relationships. On the other hand, and this is where the results derived in this paper fit, there is no simple way of testing whether a compatible parameterization of the hidden links exists. Even an attempt to rule out models that display structural incompatibility turns out to be hard when hidden variables are included.

Our results should also be viewed in the context of recent works on causal discovery, namely, the discovery of causal graphs structures containing hidden variables, which match independencies found in empirical data [VP91, SV92]. The basic assumption behind these works is that of structural stability [VP91], also called "faithfulness" [SV92], which amounts to considering all independencies as produced by the graph topology, rather than by accidental matching of numerical parameters. Our negative results no longer apply when the data is generated by such structurally stable process, and it might still be possible (though unlikely) that with such assurance, one can verify model validity (or at least I-mapness) by testing a polynomial number of independence claims. However, real-life data tends not to exhibit structural stability because its source may be from several recursive processes and may involve aggregated variables. Under such conditions, structural stability cannot be assumed and the results established in this paper, both negative and positive, provide theoretical limits on the complexity of model validation in such cases.